\documentclass{ieeeaccess}
\usepackage{cite}
\usepackage{amsmath,amssymb,amsfonts}
\usepackage{algorithmic}
\usepackage{graphicx}
\usepackage{textcomp}
\usepackage{booktabs}
\usepackage{hyperref}
\usepackage{color}
\usepackage{xcolor}
\usepackage{multirow}

\def\etal{\textit{et al.}}

\usepackage{bm}
\makeatletter
\AtBeginDocument{\DeclareMathVersion{bold}
\SetSymbolFont{operators}{bold}{T1}{times}{b}{n}
\SetSymbolFont{NewLetters}{bold}{T1}{times}{b}{it}
\SetMathAlphabet{\mathrm}{bold}{T1}{times}{b}{n}
\SetMathAlphabet{\mathit}{bold}{T1}{times}{b}{it}
\SetMathAlphabet{\mathbf}{bold}{T1}{times}{b}{n}
\SetMathAlphabet{\mathtt}{bold}{OT1}{pcr}{b}{n}
\SetSymbolFont{symbols}{bold}{OMS}{cmsy}{b}{n}
\renewcommand\boldmath{\@nomath\boldmath\mathversion{bold}}}
\makeatother

\def\BibTeX{{\rm B\kern-.05em{\sc i\kern-.025em b}\kern-.08em
    T\kern-.1667em\lower.7ex\hbox{E}\kern-.125emX}}

\begin{document}
\history{Date of publication xxxx 00, 0000, date of current version xxxx 00, 0000.}
\doi{10.1109/ACCESS.2024.0429000}

\title{Self-Supervised Multi-View Representation Learning using Vision-Language Model for 3D/4D Facial Expression Recognition}
\author{\uppercase{Muzammil Behzad}\authorrefmark{1}, \IEEEmembership{Member, IEEE}}

\address[1]{King Fahd University of Petroleum and Minerals, Saudi Arabia (e-mail: muzammil.behzad@kfupm.edu.sa)}


\markboth
{Author \headeretal: Preparation of Papers for IEEE TRANSACTIONS and JOURNALS}
{Author \headeretal: Preparation of Papers for IEEE TRANSACTIONS and JOURNALS}

\corresp{Corresponding author: Muzammil Behzad (e-mail: muzammil.behzad@kfupm.edu.sa).}

\begin{abstract}
Facial expression recognition (FER) is a fundamental task in affective computing with applications in human-computer interaction, mental health analysis, and behavioral understanding. In this paper, we propose SMILE-VLM, a self-supervised vision-language model for 3D/4D FER that unifies multiview visual representation learning with natural language supervision. SMILE-VLM learns robust, semantically aligned, and view-invariant embeddings by proposing three core components: multiview decorrelation via a Barlow Twins-style loss, vision-language contrastive alignment, and cross-modal redundancy minimization. Our framework achieves the state-of-the-art performance on multiple benchmarks. We further extend SMILE-VLM to the task of 4D micro-expression recognition (MER) to recognize the subtle affective cues. The extensive results demonstrate that SMILE-VLM not only surpasses existing unsupervised methods but also matches or exceeds supervised baselines, offering a scalable and annotation-efficient solution for expressive facial behavior understanding.
\end{abstract}


\titlepgskip=-21pt

\maketitle

\section{Introduction}

Large vision-language models (VLMs) have revolutionized the landscape of artificial intelligence by bridging the gap between visual understanding and natural language processing~\cite{bordes2024introductionvisionlanguagemodeling}. These models extend the capabilities of large language models (LLMs)~\cite{minaee2024largelanguagemodelssurvey} into the visual domain by leveraging large-scale multimodal datasets and contrastive learning objectives that enable effective joint representation learning. The success of models such as contrastive language-image pre-training (CLIP)~\cite{radford2021learningtransferablevisualmodels} has demonstrated the power of aligning images and text in a shared embedding space, enabling both zero-shot classification and cross-modal retrieval. More importantly, fine-tuning large pre-trained VLMs has also shown remarkable success in adapting to domain-specific tasks~\cite{zhai2024finetuninglargevisionlanguagemodels}, making them highly versatile across a variety of computer vision problems.

Alongside this progress, facial expression recognition (FER) has been a longstanding and crucial problem in the field of affective computing. It aims to interpret human emotions from visual facial cues, with broad applications in human-computer interaction~\cite{chowdary2023deep}, mental health monitoring~\cite{Foteinopoulou_2022}, e-learning environments~\cite{YADEGARIDEHKORDI2019103649}, and behavior analysis~\cite{7374704}. Building further on the pioneering theory of six basic emotions by Ekman and Friesen~\cite{ekman1971constants}, early FER systems predominantly relied on 2D static images and manually engineered features~\cite{LIU2023423}, which often fail to capture the subtle spatiotemporal details of facial muscle movements and generalize poorly to in-the-wild conditions.

To address these limitations, recent research has turned toward 3D and 4D FER, where the third and fourth dimensions capture depth and time, respectively. These modalities provide a richer representation of facial geometry and its temporal evolution, enabling the development of more expressive and accurate recognition systems. Within this context, a diverse range of approaches has emerged to exploit the spatiotemporal and geometric characteristics inherent in 3D facial data. Among these, local feature-based methods~\cite{6460694, 5206613, li2015efficient}, template-based techniques~\cite{4539275, 5597896}, and curve-based descriptors~\cite{samir2009intrinsic, maalej2011shape} have played a significant role in capturing local deformations and structural variations across facial regions. 

Another prominent line of work involves projection-based methods~\cite{7944639, 8265585}, which convert 3D meshes into 2D planar representations to leverage the well-established capabilities of conventional convolutional neural networks. In addition to spatial modeling, temporal dynamics have also been a critical focus. Models such as Hidden Markov Models (HMMs)~\cite{4813324, Sun:2010:TVF:1820799.1820803}, GentleBoost~\cite{sandbach2012recognition}, and random forest classifiers equipped with deformation vector fields~\cite{amor20144} have been applied to effectively capture and analyze facial motion over time. Complementary to these, spatiotemporal feature extraction techniques like local binary patterns (LBP)~\cite{FANG2012738, 6130440} and curvature-based descriptors~\cite{6553746} have demonstrated efficacy in encoding subtle expression variations across sequential 3D facial data. 

Building on these foundations, Li et al.~\cite{8373807} proposed an automatic 4D FER system using geometric images derived from differential quantities in 3D point-cloud data. Their method demonstrated the significance of score-level fusion across multiple geometric projections for robust expression prediction. These advances have highlighted the discriminative power of 3D and 4D modalities for FER. However, many of these methods still depend on supervised learning with extensive labeled datasets which is still a critical bottleneck due to the cost and subjectivity of emotion annotations.

To alleviate this dependency, recent breakthroughs in self-supervised learning (SSL) have paved the way for learning effective representations without manual labels. Notable approaches such as SimCLR~\cite{chen2020simple}, MoCo~\cite{NEURIPS2020_70feb62b}, and BYOL~\cite{NEURIPS2020_f3ada80d} leverage contrastive or predictive learning to align positive pairs while separating unrelated samples. In particular, Barlow Twins~\cite{zbontar2021barlow} introduces a decorrelation-based objective that reduces feature redundancy across positive pairs, promoting invariant yet non-collapsed embeddings. These SSL paradigms have narrowed the gap with supervised methods and opened new possibilities for learning from unlabeled 3D/4D data.

\subsection{Motivation}

Despite progress in 3D/4D FER, most existing models continue to rely on either manual feature engineering or large-scale labeled datasets. Moreover, many prior methods operate purely in the visual domain and neglect the potential of multimodal integration, particularly with natural language. With the success of VLMs in aligning visual content with textual semantics~\cite{zhai2024finetuninglargevisionlanguagemodels}, there is growing interest in incorporating language into FER systems to improve semantic understanding and generalization.

Motivated by these developments, we propose to leverage the joint strengths of self-supervised learning and vision-language modeling for 3D/4D FER. Unlike conventional 2D emotion recognition methods~\cite{8245803, 8844064, 9226082, 9369001}, which struggle to generalize across varied poses and expressions, our approach harnesses the richer spatiotemporal cues in 3D/4D facial data~\cite{li20113d, zhen2016muscular, 7163090} and aligns them with textual emotion descriptions. This combination facilitates scalable training without manual labels and supports zero-shot expression recognition. However, the limited size and complexity of available 3D/4D datasets further emphasize the need for efficient, label-agnostic learning strategies that can generalize across identities, expressions, and viewpoints.

\subsection{Role of Self-Supervision and Multimodality}

Self-supervised learning provides an attractive alternative to supervised pipelines by enabling models to learn from the structure and redundancy in the data itself. Techniques like contrastive learning, redundancy reduction, and mutual information maximization have been instrumental in extracting discriminative features from high-dimensional data without explicit labels~\cite{zbontar2021barlow}. These methods are particularly suited for multiview data, where different views of the same subject can be treated as positive pairs, while maintaining invariance to pose or lighting.

Simultaneously, multimodal vision-language learning has demonstrated impressive results in bridging low-level visual cues with high-level semantic~\cite{radford2021learningtransferablevisualmodels}. Language prompts describing emotions or affective states serve as weak supervision signals that help organize the visual representation space around human-interpretable concepts. This is particularly valuable for facial expression recognition, where the mapping between visual cues and emotional categories can be ambiguous or context-dependent. By jointly aligning multiview visual embeddings from 3D/4D facial data with corresponding language embeddings in an unsupervised setting, we propose a novel paradigm for facial expression understanding that is robust, semantically aligned, and label-efficient for 3D/4D FER.

\subsection{Contributions}

In this paper, we use CLIP~\cite{radford2021learningtransferablevisualmodels} as our baseline model to present SMILE-VLM: a Self-supervised MultI-view representation LEarning framework using Vision-Language Modeling for 3D/4D facial expression recognition. Our proposed model addresses key limitations in existing 3D/4D ER systems by reducing redundancy in SSL methods like Barlow Twins with vision-language contrastive learning to enable scalable and semantically aware representation learning. Our main contributions are summarized as follows:

\begin{itemize}
    \item Inspired by Barlow Twins~\cite{zbontar2021barlow}, we introduce a multi-view cross-correlation alignment loss to learn consistent, view-invariant, and decorrelated facial expression representations from multiple 3D views.
    \item We propose a vision-language contrastive module that aligns multiview facial embeddings with natural language descriptions of emotions, enhancing the semantic grounding of visual features and enabling zero-shot FER.
    \item We design a view-aware fusion mechanism with learnable attention weights that dynamically combine embeddings from different views based on their relative importance, improving robustness in occluded or imbalanced-view settings.
    \item We implement a cross-modal redundancy minimization objective to disentangle visual and textual modalities while retaining complementary affective features across domains.
\end{itemize}
This is worth mentioning that SMILE-VLM provides a unified, self-supervised framework for 3D/4D FER that is applicable not only to emotion recognition but also to other downstream tasks such as face recognition, anti-spoofing, and multiview identity verification.

\section{The Proposed SMILE-VLM Framework}
In this section, we describe our proposed SMILE-VLM framework: self-supervised multi-view representation learning using vision-language modeling for 3D/4D facial expression recognition. SMILE-VLM aims to leverage multi-view 3D/4D facial sequences and natural language prompts in a unified self-supervised setting. The architecture is designed to learn invariant, semantically rich, and discriminative representations without relying on explicit emotion labels. Our method is modular, scalable, and data-efficient, providing a robust pathway for developing emotion-aware systems with minimal supervision.

\subsection{Problem Formulation}
Let \( \mathcal{X} = \{x_1, x_2, \dots, x_N\} \) represent a multiview facial expression instance, where each \( x_i \) corresponds to the visual input captured from the \( i \)-th view. The different views are spatially synchronized and capture the same expression from distinct angles. In addition, we consider a natural language description \( t \in \mathcal{T} \), drawn from a predefined prompt set \( \mathcal{T} \), that semantically represents the underlying facial expression, e.g., ``a surprised face'' or ``a smiling person''. These prompts are generated using the GPT language model to map expression categories to semantically rich natural language descriptions. Some samples of the prompt templates used for generating natural language descriptions \( t \in \mathcal{T} \) are shown in Table~\ref{tab:prompt_templates}. Note that despite incorporating natural language prompts during training, SMILE-VLM remains a self-supervised learning framework. The model is not provided with categorical emotion labels. Instead, it receives semantic cues in the form of descriptive text templates that do not require manual annotation. These prompts serve as auxiliary modalities rather than ground-truth targets, guiding the model to align visual embeddings with language in a shared semantic space. Our proposed novel loss function relies entirely on unsupervised losses as explained later in subsequent sections. This design ensures that our model learns meaningful, semantically rich, and view-invariant representations without relying on any supervised classification ground-truths, making it fully self-supervised in both its formulation and training paradigm.

Our objective is to jointly learn visual and textual representations in a common embedding space. To achieve this, we define two encoders: a visual encoder \( f_v: \mathcal{X} \rightarrow \mathbb{R}^d \) and a language encoder \( f_t: \mathcal{T} \rightarrow \mathbb{R}^d \). For each input view \( x_i \in \mathcal{X} \), we generate two stochastic distortions \( \tilde{x}_i^A \) and \( \tilde{x}_i^B \), which are passed through the shared encoder and projection head to obtain the embeddings \( z_i^A = g_v(f_v(\tilde{x}_i^A)) \) and \( z_i^B = g_v(f_v(\tilde{x}_i^B)) \). These paired embeddings are used to compute the intra-view cross-correlation matrix for the multi-view embeddings.

\begin{table}[b!]
\centering
\caption{Examples of prompt templates for facial expressions.}
\label{tab:prompt_templates}
\resizebox{0.7\linewidth}{!}{%
\begin{tabular}{cc}
\toprule
\textbf{Emotion} & Template \( t \in \mathcal{T}\) \\
\toprule
Happy & 
\begin{tabular}[c]{@{}c@{}}
``a person smiling happily'',\\
``a joyful facial expression'',\\
``an expression of delight''
\end{tabular} \\\midrule
Sad & 
\begin{tabular}[c]{@{}c@{}}
``a person looking down sadly'',\\
``a face showing sorrow'',\\
``a sad expression''
\end{tabular} \\\midrule
Surprise & 
\begin{tabular}[c]{@{}c@{}}
``a surprised facial expression'',\\
``a face with surprise expression'',\\
``a face reacting with amazement''
\end{tabular} \\\midrule
Angry & 
\begin{tabular}[c]{@{}c@{}}
``a person frowning angrily'',\\
``an expression of frustration'',\\
``a face showing intense anger''
\end{tabular} \\\midrule
Disgust & 
\begin{tabular}[c]{@{}c@{}}
``a person showing disgust'',\\
``a face with disgust expression'',\\
``a disgusted facial reaction''
\end{tabular} \\\midrule
Fear & 
\begin{tabular}[c]{@{}c@{}}
``a fearful facial expression'',\\
``a person appearing afraid'',\\
``a face with fear expression''
\end{tabular} \\
\bottomrule
\end{tabular}%
}
\end{table}

\subsection{Model Overview}
SMILE-VLM is composed of three primary components: a multiview visual encoder, a text encoder, and a series of loss functions that enforce inter-view consistency, vision-language alignment, and cross-modal redundancy minimization. Each visual stream passes through an encoder backbone and a nonlinear projector, generating view-specific embeddings \( \{z_1', z_2', ..., z_N'\} \). These embeddings are later fused using a view-aware attention mechanism to yield an aggregated representation \( z^{mv} \). In parallel, text prompts are mapped to the shared space using a language encoder resulting in the embedding \( z^t \).

\begin{figure*}[t!]
    \centering
    \includegraphics[width=\linewidth]{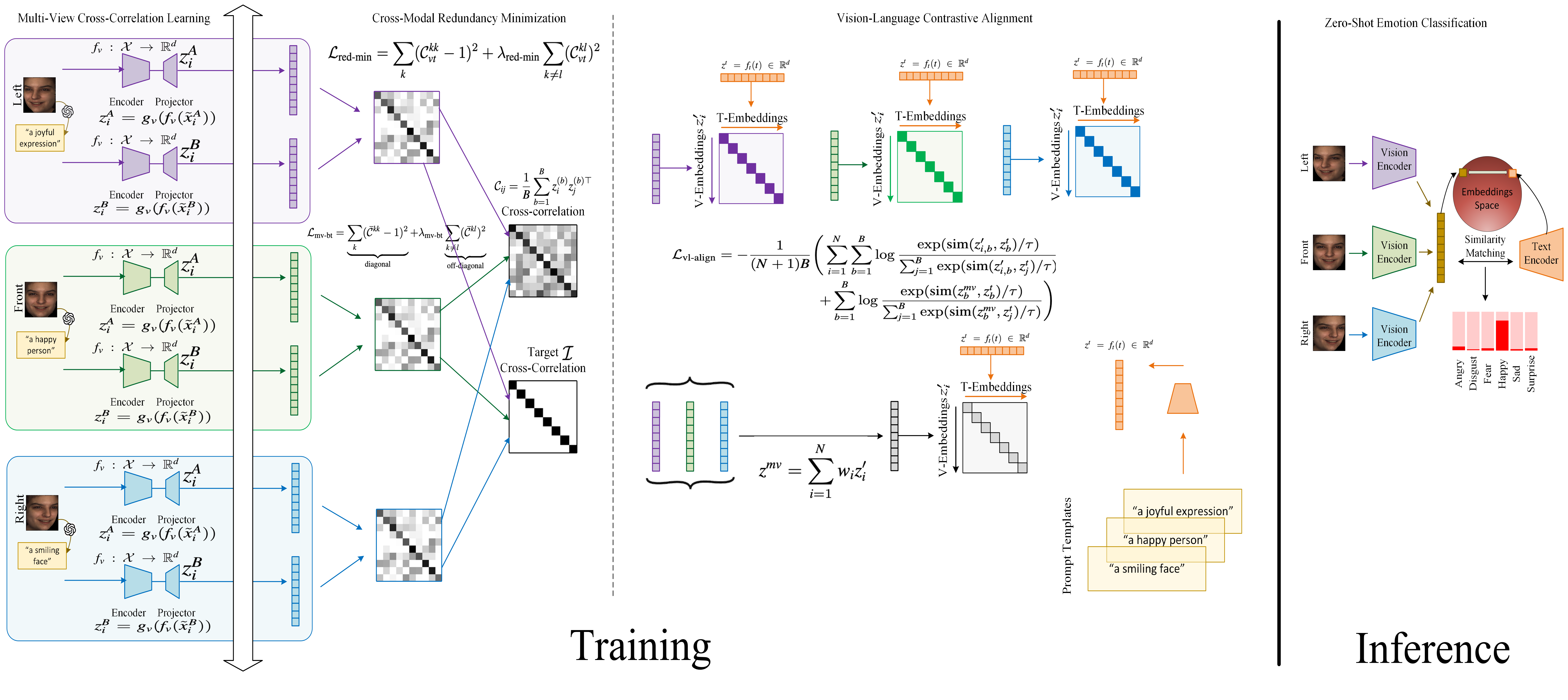}
    \caption{Overview of the proposed SMILE-VLM training pipeline. The multi-view facial inputs are encoded and projected into a joint embedding space, combined using view-aware fusion, and aligned with textual descriptions via vision-language contrastive learning. These embeddings are then used to optimize three key objectives. First, a multiview cross-correlation loss encourages the visual representations to be both invariant across views and decorrelated across feature dimensions. Second, a vision-language alignment loss brings the fused multiview visual embedding and individual view embeddings close to their corresponding textual descriptions. Third, a cross-modal redundancy reduction loss minimizes redundant information between visual and textual modalities, ensuring that each contributes complementary features to the shared embedding space.}
    \label{fig:mainmodelFINAL}
\end{figure*}

These embeddings are used to optimize three key objectives: (1) a multiview cross-correlation loss that encourages invariant and decorrelated view representations; (2) a vision-language alignment loss that brings the fused visual embedding close to its textual description; and (3) a cross-modal redundancy reduction loss that enforces complementarity between visual and linguistic features. An overview of the proposed SMILE-VLM model is shown in Fig. \ref{fig:mainmodelFINAL}.

\subsection{Multi-View Cross-Correlation Learning}
The SMILE-VLM framework generalizes the Barlow Twins objective to a multi-view and distortion-aware setting, ensuring that learned visual representations are both invariant to viewpoint and non-redundant across feature dimensions. To achieve this, we introduce a two-stage strategy that first computes per-view cross-correlations based on augmentations and then aggregates them into a unified decorrelation target. For each input view \( x_i \), we generate two stochastic augmentations, denoted \( \tilde{x}_i^A \) and \( \tilde{x}_i^B \), simulating different distortions of the same input. These augmented views are encoded and projected into corresponding embeddings \( z_i^A = g_v(f_v(\tilde{x}_i^A)) \) and \( z_i^B = g_v(f_v(\tilde{x}_i^B)) \). A cross-correlation matrix \( \mathcal{C}_i \) is computed for each view individually as:
\begin{equation}
    \mathcal{C}_i = \frac{1}{B} \sum_{b=1}^{B} z_i^{A(b)} z_i^{B(b)\top},
\end{equation}
where \( B \) is the batch size, and \( z_i^{A(b)}, z_i^{B(b)} \) are the embeddings for the \( b \)-th sample in augmentations A and B, respectively. To promote global consistency, we average the individual correlation matrices to form a unified target matrix as given below:
\begin{equation}
    \bar{\mathcal{C}} = \frac{1}{N} \sum_{i=1}^N \mathcal{C}_i,
\end{equation}
where \( N \) is the number of views. This averaged matrix captures a shared structural representation that incorporates information from all spatial perspectives. We then define the multi-view Barlow Twins loss as a composite objective composed of two terms:
\begin{equation}
\mathcal{L}_{\text{mv-bt}} = \sum_{k} \left( \left( \frac{1}{N} \sum_{i=1}^{N} \mathcal{C}_i^{kk} \right) - 1 \right)^2 + \lambda \sum_{k \ne l} \left( \frac{1}{N} \sum_{i=1}^{N} \mathcal{C}_i^{kl} \right)^2.
\end{equation}
The above expression can be represented more compactly as:
\begin{equation}
    \mathcal{L}_{\text{mv-bt}} = \underbrace{\sum_{k} (\bar{\mathcal{C}}^{kk} - 1)^2}_{\text{diagonal}} + \lambda_{\text{mv-bt}} \underbrace{\sum_{k \ne l} (\bar{\mathcal{C}}^{kl})^2}_{\text{off-diagonal}},
\end{equation}
where the diagonal term encourages the self-correlation between dimensions to be close to 1 (indicating high variance and information preservation), and the off-diagonal term penalizes redundancy between feature dimensions. The hyperparameter \( \lambda_{\text{mv-bt}} \) balances the contribution of the two terms.

This formulation has two primary advantages. First, it ensures that the model learns robust view-invariant features by collapsing embeddings from augmented views of the same instance while preserving diversity across dimensions. Second, it avoids the risk of representation collapse by maximizing variance and minimizing redundancy in the learned space. By averaging across all views, the model also ensures that no single perspective dominates, resulting in balanced and consistent multi-view feature alignment. Overall, this loss effectively enforces spatial and dimensional decorrelation, which is critical for self-supervised learning in multiview 3D/4D FER.

\subsection{Vision-Language Contrastive Alignment}
To incorporate semantic understanding into the representation learning process, we extend the vision-language alignment module in SMILE-VLM to operate across both individual views and the fused multiview representation. This design enables the model to learn semantically meaningful representations from all available visual perspectives, as well as from their joint integration, thereby strengthening both generalization and interpretability.

Each image view \( x_i \in \mathcal{X} \) is encoded into a projected visual embedding \( z_i' = g_v(f_v(x_i)) \in \mathbb{R}^d \), and each sample's corresponding text prompt \( t \in \mathcal{T} \) is encoded using the text encoder to obtain a textual embedding \( z^t = f_t(t) \in \mathbb{R}^d \). A dynamic fusion module then combines the view-specific embeddings into a unified multiview embedding using learned attention weights:
\begin{equation}
    z^{mv} = \sum_{i=1}^N w_i z_i', \quad \text{where } \sum_i w_i = 1, \; w_i \geq 0.
    \label{eq:VLCA}
\end{equation}

To ensure comprehensive semantic alignment, we compute a contrastive InfoNCE loss \cite{oord2019representationlearningcontrastivepredictive} not only between the fused embedding and its associated textual description, but also between each individual view embedding and the same text. This results in \( N+1 \) alignment computations per sample in each batch. The extended vision-language alignment loss is defined as:
\begin{equation}
\begin{aligned}
\mathcal{L}_{\text{vl-align}} = -\frac{1}{(N+1)B} \Bigg( \sum_{i=1}^{N} \sum_{b=1}^B \log \frac{\exp(\text{sim}(z_{i,b}', z^t_b)/\tau)}{\sum_{j=1}^B \exp(\text{sim}(z_{i,b}', z^t_j)/\tau)} \\ 
+ \sum_{b=1}^B \log \frac{\exp(\text{sim}(z^{mv}_b, z^t_b)/\tau)}{\sum_{j=1}^B \exp(\text{sim}(z^{mv}_b, z^t_j)/\tau)} \Bigg),
\end{aligned}
\end{equation}
where \( \text{sim}(a, b) = \frac{a^\top b}{\|a\| \|b\|} \) denotes cosine similarity, \( \tau \) is a temperature hyperparameter, and \( B \) is the batch size.

This formulation benefits from both cross-modal diversity and view-specific details. By explicitly aligning each view to the textual description, the model becomes more sensitive to view-dependent understanding in expression. The fusion alignment further reinforces a consistent semantic anchor in the shared embedding space. This dual-level contrastive supervision encourages richer modality interaction and leads to robust representations that generalize well to unseen expressions. Additionally, the alignment loss incorporates hard negatives within the batch, which further sharpens the separation between similar but semantically distinct expressions, ultimately enhancing the discriminative power of the learned features.

\subsection{View-Aware Embedding Fusion}

The uniform aggregation of view embeddings may discard discriminative cues from the most informative facial views. To address this, we propose a view-aware fusion module that assigns dynamic importance scores to each view as expressed in Eq. (\ref{eq:VLCA}). Each view embedding \( z_i' \) is pooled with global average pooling and passed through a lightweight MLP to produce a scalar score \( s_i \). The fusion weights are then computed via a softmax operation as:
\begin{equation}
    s_i = \text{MLP}(\text{pool}(z_i')), \quad w_i = \frac{\exp(s_i)}{\sum_{j=1}^N \exp(s_j)}.
\end{equation}
This mechanism enables the model to attend more heavily to views that contribute maximally to expression discriminability, improving both robustness and performance.

\subsection{Cross-Modal Redundancy Minimization}
The excessive correlation between visual and textual features may reduce their significance ultimately impacting the model negatively. To ensure a proper and valid vision-language alignment, we introduce a redundancy minimization term that penalizes off-diagonal entries of the visual-textual correlation matrix. Given visual-text pairs \( (z^{mv}_b, z^t_b) \), we define their batch-wise correlation as:
\begin{equation}
    \mathcal{C}_{vt} = \frac{1}{B} \sum_{b=1}^{B} z^{mv}_b z^t_b{}^\top.
\end{equation}
This matrix quantifies the dimension-wise correlation between modalities across the entire batch. We then define the redundancy minimization loss as follows:
\begin{equation}
    \mathcal{L}_{\text{red-min}} = \sum_k (\mathcal{C}_{vt}^{kk} - 1)^2 + \lambda_{\text{red-min}} \sum_{k \ne l} (\mathcal{C}_{vt}^{kl})^2,
\end{equation}
where the first term enforces the diagonal elements (self-correlations) to approach 1, ensuring each dimension retains variance across modalities. The second term penalizes off-diagonal entries, which represent undesirable correlations between different feature dimensions. The hyperparameter \( \lambda_{\text{red-min}} \) balances the influence of variance maximization and redundancy suppression. This loss encourages disentanglement between modalities, ensuring that visual and textual features encode distinct yet complementary information. As a result, the learned embeddings retain richer semantics and improve robustness in downstream tasks such as zero-shot emotion classification. By reducing redundancy, this term reinforces the benefits of vision-language fusion in a self-supervised setting without reliance on ground-truth labels. 

\subsection{Joint Loss Optimization for Cross-Modal Learning}

The joint loss optimization for SMILE-VLM integrates all proposed loss functions into a unified formulation as:
\begin{equation}
    \mathcal{L}_{\text{SMILE-VLM}} = \alpha \mathcal{L}_{\text{mv-bt}} + \beta \mathcal{L}_{\text{vl-align}} + \gamma \mathcal{L}_{\text{red-min}},
\end{equation}
where \( \alpha, \beta, \gamma \) are weighting coefficients that control the relative influence of each loss term. These weights are critical in balancing the learning dynamics of the model, particularly when jointly optimizing loss formulations with slightly different gradients and convergence behaviors.

The coefficient \( \alpha \) refers to the strength of the multi-view Barlow Twins loss, which promotes consistent and decorrelated representations across facial views. A higher value of \( \alpha \) prioritizes view-invariant representation learning, which is especially beneficial when dealing with substantial inter-view variations. However, overly dominant \( \alpha \) may result in under-utilization of semantic guidance from language descriptions. The coefficient \( \beta \) determines the emphasis on the vision-language alignment loss. A moderate to strong \( \beta \) encourages the model to anchor visual features to semantically rich textual prompts, improving generalization and zero-shot capabilities. If \( \beta \) is too low, the learned features may remain visually aligned but semantically shallow, limiting interpretability. The coefficient \( \gamma \) controls the contribution of the cross-modal redundancy minimization loss. This component enforces disentanglement between modalities, reducing redundancy and enhancing the capacity of the joint embedding space. A carefully tuned \( \gamma \) helps prevent overfitting to shared cues and promotes distinctive learning between visual and textual domains. Overall, these hyperparameters must be chosen to reflect task-specific priorities. For example, in resource-constrained or few-shot settings, placing more weight on \( \beta \) may aid in leveraging pretrained language knowledge. On the other hand, robust view-invariance learning might necessitate prioritizing \( \alpha \).

\section{Experimental Setup}
\subsection{Datasets}
We evaluate and validate the performance of the proposed SMILE-VLM model using four widely recognized benchmark datasets: Bosphorus~\cite{savran2008bosphorus}, BU-3DFE~\cite{yin20063d}, BU-4DFE~\cite{4813324}, and BP4D-Spontaneous~\cite{ZHANG2014692}. These datasets offer a comprehensive range of facial expressions and subject variations, encompassing both posed and spontaneous affective behaviors in 3D and 4D point-cloud formats. Bosphorus and BU-3DFE provide detailed static 3D scans under controlled conditions, while BU-4DFE and BP4D deliver dynamic sequences that capture temporal evolution of expressions. This diversity allows thorough evaluation of both spatial and temporal aspects of facial expression recognition. 

\subsection{Preprocessing and View Selection}

Following standard evaluation protocols from previous works~\cite{8373807, 8023848, 9320291, behzad2021Sparse3D, behzad2021disentangling}, we convert raw 3D and 4D point-cloud data into multiview 2D projections. For each frame or mesh, we generate views at frontal (0$^\circ$), left ($-30^\circ$), and right ($+30^\circ$) angles to simulate realistic camera perspectives. In the case of 4D sequences, temporal frames are uniformly sampled and converted into compact dynamic image representations using rank pooling~\cite{bilen2018action}, which preserves temporal motion patterns while reducing computational complexity.

\subsection{Language Prompt Engineering}

To enable multimodal alignment while maintaining self-supervised learning, we generate expression-related text prompts using the GPT language model. Each expression category is associated with a set of semantically rich prompts (e.g., ``a joyful smile'', ``a person smiling happily'', ``a delighted facial expression''). These descriptions are randomly sampled at training time to provide diversity and prevent overfitting. All prompts are encoded using the model's text encoder to produce fixed-length embeddings that act as soft, descriptive anchors in the joint representation space, without acting as hard labels.

\subsection{Implementation Details}
The textual features are extracted from the frozen CLIP text encoder. The model is trained using the Adam optimizer using an initial learning rate of 1e-4 with a weight decay factor. The training and inference are carried out using PyTorch on distributed NVIDIA GeForce RTX 3090 Ti GPUs. Once the model is trained, the inference is done by passing the multiview facial inputs through the visual encoders and projections head to obtain view-level embeddings. These embeddings are fused using the learned weights to produce a unified visual representation. The standard classification is done in a zero-shot setting, where the fused embedding of a query sample is matched directly against the encoded textual prompts. Finally, the class with the highest similarity score is selected as the predicted expression.

\section{Results and Analysis}
To the best of our knowledge, only one prior method has explored 3D/4D facial expression recognition in a fully \color{magenta}unsupervised \color{black} setting~\cite{behzad2021self}. We include this method in our evaluation to establish a direct baseline for self-supervised learning in this domain. In addition, we compare SMILE-VLM against several state-of-the-art \color{teal}supervised \color{black} approaches to provide a broad assessment of our model's performance. To ensure reliable and generalizable evaluation, we adopt a 10-fold subject-independent cross-validation protocol across all datasets. This strategy guarantees that no subject appears in both the training and testing sets, effectively removing identity-specific leakage and ensuring that models are evaluated under strict generalization conditions. Such a protocol is essential in affective computing tasks, where subject overlap can lead to inflated metrics and poor deployment robustness.

\subsection{Performance on 3D FER}
Following established evaluation protocols in prior studies~\cite{7944639, 8265585}, we conduct experiments on the BU-3DFE and Bosphorus datasets to evaluate the effectiveness of our proposed unsupervised model for 3D facial expression recognition. The BU-3DFE dataset includes 101 subjects and is typically partitioned into two subsets: Subset I, comprising samples with expressions at the two highest intensity levels and widely used as the standard benchmark; and Subset II, which includes expressions across all four intensity levels but excludes 100 neutral scans and is less frequently used in prior 3D FER research. For the Bosphorus dataset, we follow the common practice of selecting only the 65 subjects who performed all six basic facial expressions, ensuring consistency with prior evaluation settings.
\begin{table}[b!]
	\caption{Comparison of accuracy (\%) with state-of-the-art methods on the BU-3DFE Subset I, Subset II, and Bosphorus datasets.}
	\label{table:3DFERresults}
	\resizebox{\linewidth}{!}{%
		\begin{tabular}{l c}
			\hline
			Method & Subset I  (\color{blue}$\uparrow$\color{red}$\downarrow$\color{black})\\
			\hline			
			{\color{teal}Zhen \etal \cite{zhen2016muscular}} & 84.50 (\color{blue}5.01$\uparrow$\color{black}) \\ 
			{\color{teal}Yang \etal \cite{7163090}} & 84.80 (\color{blue}4.71$\uparrow$\color{black}) \\
			{\color{teal}Li \etal \cite{li2015efficient}} & 86.32 (\color{blue}3.19$\uparrow$\color{black}) \\
			{\color{teal}Li \etal \cite{7944639}} & 86.86 (\color{blue}2.65$\uparrow$\color{black}) \\ 
			{\color{teal}Oyedotun \etal \cite{8265585}} & 89.31  (\color{blue}0.20$\uparrow$\color{black}) \\ 
			\hline
			{\color{magenta}MiFaR \cite{behzad2021self}} & 88.53 (\color{blue}0.98$\uparrow$\color{black})\\ 
			\textbf{{\color{magenta}SMILE-VLM (Ours)}} & \textbf{89.51} \\ 
			\hline
		\end{tabular} 
		\begin{tabular}{l c c}
			\hline
			Method & Subset II (\color{blue}$\uparrow$\color{red}$\downarrow$\color{black}) & Bosphorus  (\color{blue}$\uparrow$\color{red}$\downarrow$\color{black})\\
			\hline
			{\color{teal}Li \etal \cite{li2015efficient}} & 80.42 (\color{blue}3.59$\uparrow$) & 79.72 (\color{blue}0.25$\uparrow$\color{black})\\ 
			{\color{teal}Yang \etal \cite{7163090}} & 80.46 (\color{blue}3.55$\uparrow$) & 77.50 (\color{blue}2.47$\uparrow$\color{black})\\ 
			{\color{teal}Li \etal \cite{7944639}} & 81.33 (\color{blue}2.68$\uparrow$) & 80.00 (\color{red}0.03$\downarrow$\color{black})\\ 
			\hline
			{\color{magenta}MiFaR \cite{behzad2021self}} & 82.67 (\color{blue}1.34$\uparrow$\color{black}) & 78.84 (\color{blue}1.13$\uparrow$\color{black}) \\ 
			\textbf{{\color{magenta}SMILE-VLM (Ours)}} & \textbf{84.01} & \textbf{79.97} \\ 
			\hline
		\end{tabular}
        }
\end{table}

In Table~\ref{table:3DFERresults}, our model demonstrates competitive performance across multiple 3D facial expression recognition benchmarks. On Subset I of the BU-3DFE dataset, the proposed SMILE-VLM model achieves an accuracy of 89.51\%, slightly surpassing the best-performing supervised method~\cite{8265585}, with a small improvement of \color{blue}0.20\%\color{black}. More notably, our model outperforms the prior unsupervised method MiFaR~\cite{behzad2021self} by a margin of \color{blue}0.98\%\color{black}, underscoring the effectiveness of our multi-view vision-language self-supervised learning framework. On Subset II of the BU-3DFE dataset, SMILE-VLM sets a new state-of-the-art with an accuracy of 84.01\%, outperforming both supervised and unsupervised baselines. In particular, it improves upon MiFaR~\cite{behzad2021self} by \color{blue}1.34\% \color{black} and exceeds the performance of the supervised method by Li \etal~\cite{7944639} by \color{blue}2.68\%\color{black}. This result is especially significant as Subset II includes more varied expression intensity levels, making it a more challenging benchmark.

Similarly, on the Bosphorus dataset, SMILE-VLM achieves an accuracy of 79.97\%, demonstrating robust generalization to diverse 3D expression data. It outperforms the unsupervised MiFaR baseline by \color{blue}1.13\% \color{black} and competes closely with top-performing supervised models, while maintaining a fully unsupervised training setup. These results collectively demonstrate that SMILE-VLM delivers performance on par with or superior to leading supervised methods in 3D facial expression recognition.

\subsection{Performance on 4D FER}
To evaluate the effectiveness of our proposed model on 4D FER, we conducted comprehensive experiments on the BU-4DFE dataset, which consists of 3D video sequences of 101 subjects performing six posed facial expressions. Table~\ref{table:4DFERresults} presents the performance comparison with the state-of-the-art methods under similar experimental settings. Our model achieves the highest accuracy of 96.57\%, surpassing both supervised and unsupervised methods, attributed to our joint multiview and vision-language learning framework. In particular, our model outperforms the traditional supervised method by Zhen \etal~\cite{8023848} by a margin of \color{blue}1.44\%\color{black}, and shows clear advantages over methods using key-frame selection or sliding window strategies. These consistent improvements emphasize the ability of our architecture to effectively capture the spatiotemporal dynamics inherent in 4D facial expressions.

Additionally, compared to the only existing unsupervised baseline MiFaR~\cite{behzad2021self}, which achieved 95.76\%, SMILE-VLM achieves a relative gain of \color{blue}0.81\%\color{black}, setting a new benchmark for unsupervised 4D FER. While prior supervised methods depend heavily on labeled data, our approach attains comparable or even superior performance without requiring manual annotations. This significantly reduces annotation costs and improves scalability for real-world deployment. The competitive performance of SMILE-VLM over both supervised and unsupervised baselines underscores the strength of our joint multiview learning formulation in capturing expressive facial behaviors over time.

\begin{table}[t!]
	\caption{Comparison of 4D facial expression recognition performance (\%) with state-of-the-art methods on the BU-4DFE dataset.}
	\label{table:4DFERresults}
	\begin{center}
            \resizebox{\columnwidth}{!}{ 
		\begin{tabular}{l c c}
			\hline
			Method & Experimental Settings & Accuracy (\color{blue}$\uparrow$\color{red}$\downarrow$\color{black}) \\
			\hline
			{\color{teal}Sandbach \etal \cite{sandbach2012recognition}} & 6-CV, Sliding window & 64.60 ({\color{blue}31.97$\uparrow$})\\ 
			{\color{teal}Fang \etal \cite{6130440}} & 10-CV, Full sequence & 75.82 ({\color{blue}20.75$\uparrow$})\\ 
			{\color{teal}Xue \etal \cite{7045888}} & 10-CV, Full sequence & 78.80 ({\color{blue}17.77$\uparrow$})\\ 
			{\color{teal}Sun \etal \cite{Sun:2010:TVF:1820799.1820803}} & 10-CV, - & 83.70 ({\color{blue}12.87$\uparrow$})\\
			{\color{teal}Zhen \etal \cite{7457243}} & 10-CV, Full sequence & 87.06 ({\color{blue}9.51$\uparrow$})\\ 
			{\color{teal}Yao \etal \cite{10.1145/3131345}} & 10-CV, Key-frame & 87.61 ({\color{blue}8.96$\uparrow$})\\ 
			{\color{teal}Fang \etal \cite{FANG2012738}} & 10-CV, - & 91.00 ({\color{blue}5.57$\uparrow$})\\ 
			{\color{teal}Li \etal \cite{8373807}} & 10-CV, Full sequence & 92.22 ({\color{blue}4.35$\uparrow$})\\ 
			{\color{teal}Ben Amor \etal \cite{amor20144}} & 10-CV, Full sequence & 93.21 ({\color{blue}3.36$\uparrow$})\\ 
			{\color{teal}Zhen \etal \cite{8023848}} & 10-CV, Full sequence & 94.18 ({\color{blue}2.39$\uparrow$})\\ 
			{\color{teal}Bejaoui \etal \cite{Bejaoui2019}} & 10-CV, Full sequence & 94.20 ({\color{blue}2.37$\uparrow$})\\ 
			{\color{teal}Zhen \etal \cite{8023848}} & 10-CV, Key-frame & 95.13 ({\color{blue}1.44$\uparrow$})\\ 
			{\color{teal}Behzad \etal \cite{behzad2019automatic}}  & 10-CV, Full sequence & 96.50 ({\color{blue}0.07$\uparrow$})\\ 
			\hline
			{\color{magenta}MiFaR \cite{behzad2021self}} & 10-CV, Full sequence & 95.76 ({\color{blue}0.81$\uparrow$})\\ 
			\textbf{{\color{magenta}SMILE-VLM (Ours)}} & 10-CV, Full sequence & \textbf{96.57}\\ 
			\hline
		\end{tabular}
             } 
	\end{center}
\end{table}
\begin{table}[b!]
	\caption{Comparison of recognition accuracy (\%) with state-of-the-art methods on the BP4D-Spontaneous dataset.\vspace{0.1cm} \hspace{\textwidth} \hspace*{1.2cm}(a) Recognition \hspace{1cm} (b) Cross-Dataset Evaluation\vspace{-0.15cm}}
	\label{table:4DFERresults_part2}
	\resizebox{\linewidth}{!}{%
		\begin{tabular}{l c}
			\hline
			Method & Accuracy (\color{blue}$\uparrow$\color{red}$\downarrow$\color{black})\\
			\hline
			{\color{teal}Yao \etal \cite{10.1145/3131345}} & 86.59 (\color{blue}1.86$\uparrow$\color{black})\\ 
			{\color{teal}Danelakis \etal \cite{danelakis2016effective}} & 88.56 (\color{red}0.11$\downarrow$\color{black})\\ 
			\hline
            \color{magenta}MiFaR \cite{behzad2021self} & 87.14 ({\color{blue}1.31$\uparrow$})\\ 
            \textbf{{\color{magenta}SMILE-VLM (Ours)}} & \textbf{88.45}\\ 
			\hline
		\end{tabular}
		\begin{tabular}{l c}
			\hline
			Method & Accuracy (\color{blue}$\uparrow$\color{red}$\downarrow$\color{black})\\
			\hline
			{\color{teal}Zhang \etal \cite{ZHANG2014692}} & 71.00 (\color{blue}9.66$\uparrow$\color{black})\\ 
			{\color{teal}Zhen \etal \cite{zhen2017magnifying}} & 81.70 (\color{red}1.04$\downarrow$\color{black})\\ 
			\hline
            \color{magenta}MiFaR \cite{behzad2021self} & 79.05 ({\color{blue}1.61$\uparrow$})\\ 
            \textbf{{\color{magenta}SMILE-VLM (Ours)}} & \textbf{80.66}\\ 
			\hline
		\end{tabular}
	}
\end{table}
\subsection{Towards Spontaneous 4D FER}
To validate our model's capability in recognizing spontaneous expressions, we conduct experiments on the BP4D-Spontaneous dataset, which contains 41 subjects displaying natural facial responses, including additional emotion categories such as nervousness and pain. We summarize our results for both recognition and cross-dataset evaluation tasks in Table~\ref{table:4DFERresults_part2}. In the recognition setting, our proposed SMILE-VLM model achieves the highest unsupervised accuracy of 88.45\%, outperforming the prior unsupervised method MiFaR~\cite{behzad2021self} (87.14\%) by a margin of \color{blue}1.31\%\color{black}, and improving over the supervised state-of-the-art method by Yao \etal~\cite{10.1145/3131345} (86.59\%) by \color{blue}1.86\%\color{black}. While our model slightly trails behind the fully supervised approach of Danelakis \etal~\cite{danelakis2016effective}, which achieved 88.56\%, the difference is marginal at only \color{red}0.11\%\color{black}. These results reinforce the effectiveness of our joint multiview and self-supervised representation learning framework in handling spontaneous, naturally occurring facial expressions without reliance on labeled data.

To further test the robustness and generalizability of our model, we adopt a cross-dataset evaluation protocol, following established practices in the literature~\cite{ZHANG2014692, zhen2017magnifying}. Specifically, we train SMILE-VLM on the BU-4DFE dataset and evaluate it on the BP4D-Spontaneous dataset, focusing on Tasks 1 and 8, which correspond to happy and disgust expressions. This setting is particularly valuable for assessing how well a model generalizes across datasets with different subject identities, and emotional distributions. We show that our model achieves an accuracy of 80.66\% in this setup, outperforming MiFaR~\cite{behzad2021self} by \color{blue}1.61\% \color{black} (79.05\%), and demonstrating a substantial improvement over the earlier supervised method by Zhang \etal~\cite{ZHANG2014692} (71.00\%) by \color{blue}9.66\%\color{black}. While SMILE-VLM trails the supervised method of Zhen \etal~\cite{zhen2017magnifying} (81.70\%) by a small margin of \color{red}1.04\%\color{black}, it is important to emphasize that our approach operates entirely in an unsupervised manner. These findings confirm that SMILE-VLM provides strong generalization capabilities, bridging the gap with supervised methods, and offering a scalable solution for facial expression recognition in spontaneous and real-world scenarios.

\subsection{Ablation Study}
\subsubsection{Effectiveness of Each Component in SMILE-VLM}
\begin{figure}[t!]
    \centering
	\includegraphics[width=\linewidth]{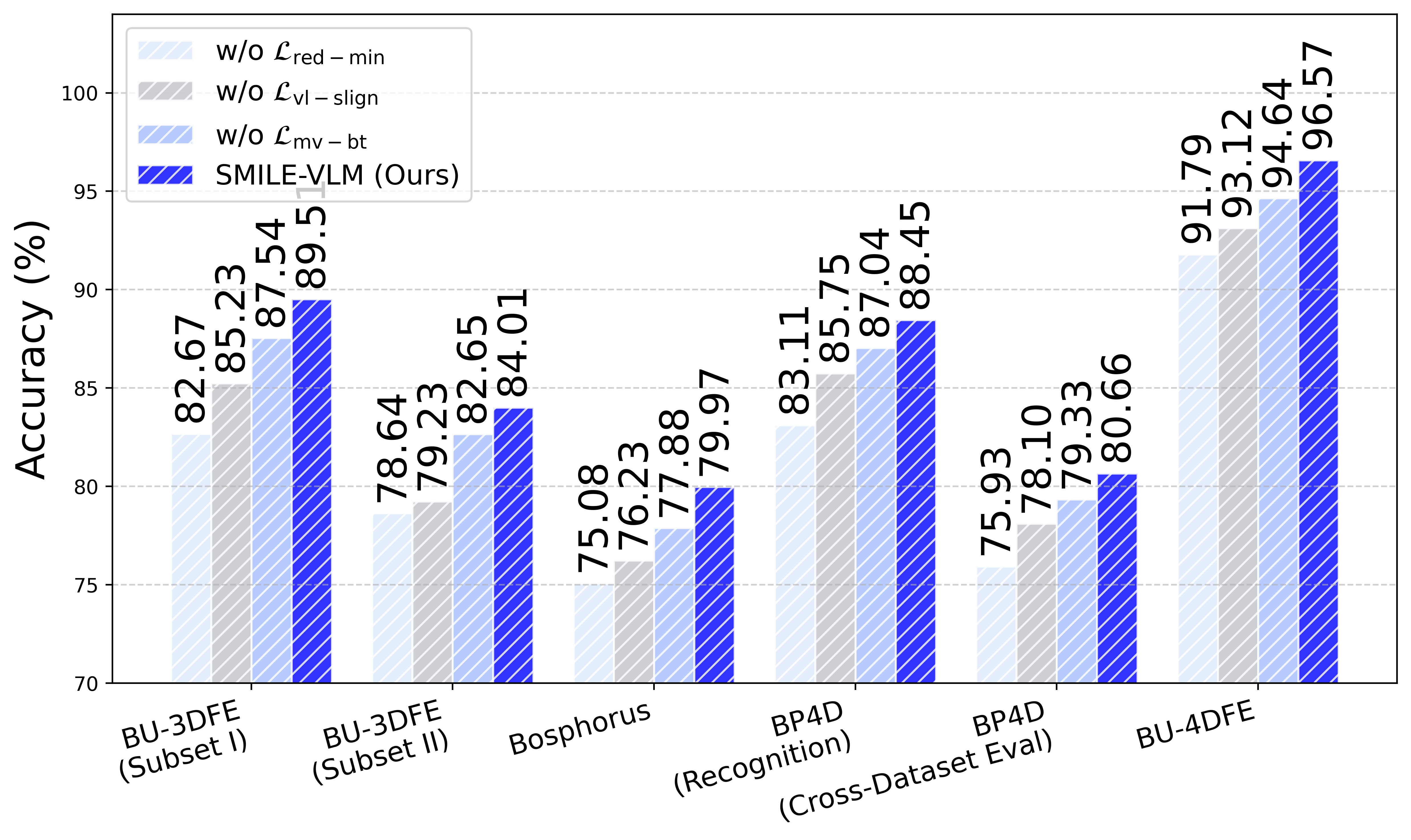}
	\caption{Ablation study of SMILE-VLM on multiple datasets.}
	\label{fig:ablation}
\end{figure}

To assess the contribution of each key component in the SMILE-VLM framework, we perform a comprehensive ablation study across six benchmark settings, as depicted in Fig.~\ref{fig:ablation}. Specifically, we evaluate the impact of removing each of the three major components: the redundancy minimization loss \( \mathcal{L}_{\text{red-min}} \), the vision-language alignment loss \( \mathcal{L}_{\text{vl-align}} \), and the multi-view Barlow Twins loss \( \mathcal{L}_{\text{mv-bt}} \). The results show that removing any one of these losses leads to a consistent drop in performance across all datasets. Notably, the absence of \( \mathcal{L}_{\text{red-min}} \) results in the largest degradation on BU-3DFE Subset I and BP4D Recognition, underscoring its role in reducing cross-modal redundancy. The absence of \( \mathcal{L}_{\text{vl-align}} \) primarily affects cross-dataset generalization (e.g., BP4D cross-dataset evaluation), where language-guided semantic consistency is crucial. Meanwhile, dropping \( \mathcal{L}_{\text{mv-bt}} \) significantly lowers performance in multiview-sensitive datasets like BU-3DFE and BU-4DFE. In contrast, the full SMILE-VLM model consistently achieves the highest accuracy across all benchmarks, confirming that each module plays a significant role. This clearly demonstrates the effectiveness of our proposed multiview and vision-language integration framework.

\begin{table*}[h]
\centering
\caption{Comparison of ME Emotion Recognition Performance on the 4DME dataset.}
\label{tab:4dme_mer}
\resizebox{\linewidth}{!}{%
\begin{tabular}{llcccccc}
\toprule
\textbf{Metric} & \textbf{Model/Profiles} & \textbf{Positive} & \textbf{Negative} & \textbf{Surprise} & \textbf{Repression} & \textbf{Others} & \textbf{Average} \\
\midrule
\multirow{5}{*}{F1-score}
& Left~\cite{li20224dme} & 0.5971 & 0.6639 & 0.6040 & 0.5398 & 0.5804 & 0.5970 \\
& Right~\cite{li20224dme} & 0.5249 & 0.6601 & 0.5900 & 0.5404 & 0.5739 & 0.5778 \\
& Front~\cite{li20224dme} & 0.6367 & 0.6766 & 0.6313 & 0.7059 & 0.7298 & 0.6760 \\
& Multi-views~\cite{li20224dme} & 0.7443 & 0.8347 & 0.8034 & 0.7966 & 0.7750 & 0.7908 \\
\cmidrule(lr){2-8}
& SMILE-VLM (multi-views) & \textbf{0.7612} & \textbf{0.8458} & \textbf{0.8126} & \textbf{0.8093} & \textbf{0.7824} & \textbf{0.8023} \\
\midrule
\multirow{5}{*}{Accuracy (\%)}
& Left~\cite{li20224dme} & 66.10 & 66.53 & 66.95 & 65.68 & 69.07 & 66.86 \\
& Right~\cite{li20224dme} & 61.02 & 66.10 & 64.83 & 66.53 & 68.22 & 65.34 \\
& Front~\cite{li20224dme} & 69.07 & 68.22 & 67.80 & 82.63 & 83.90 & 74.32 \\
& Multi-views~\cite{li20224dme} & 80.08 & 83.47 & 85.59 & 91.10 & 87.71 & 85.59 \\
\cmidrule(lr){2-8}
& SMILE-VLM (multi-views) & \textbf{81.62} & \textbf{84.60} & \textbf{86.22} & \textbf{92.18} & \textbf{88.44} & \textbf{86.61} \\
\bottomrule
\end{tabular}
}
\end{table*}

\subsubsection{Accuracy Improvements Across Datasets} 
\begin{figure*}[h!]
    \centering
	\includegraphics[width=\linewidth]{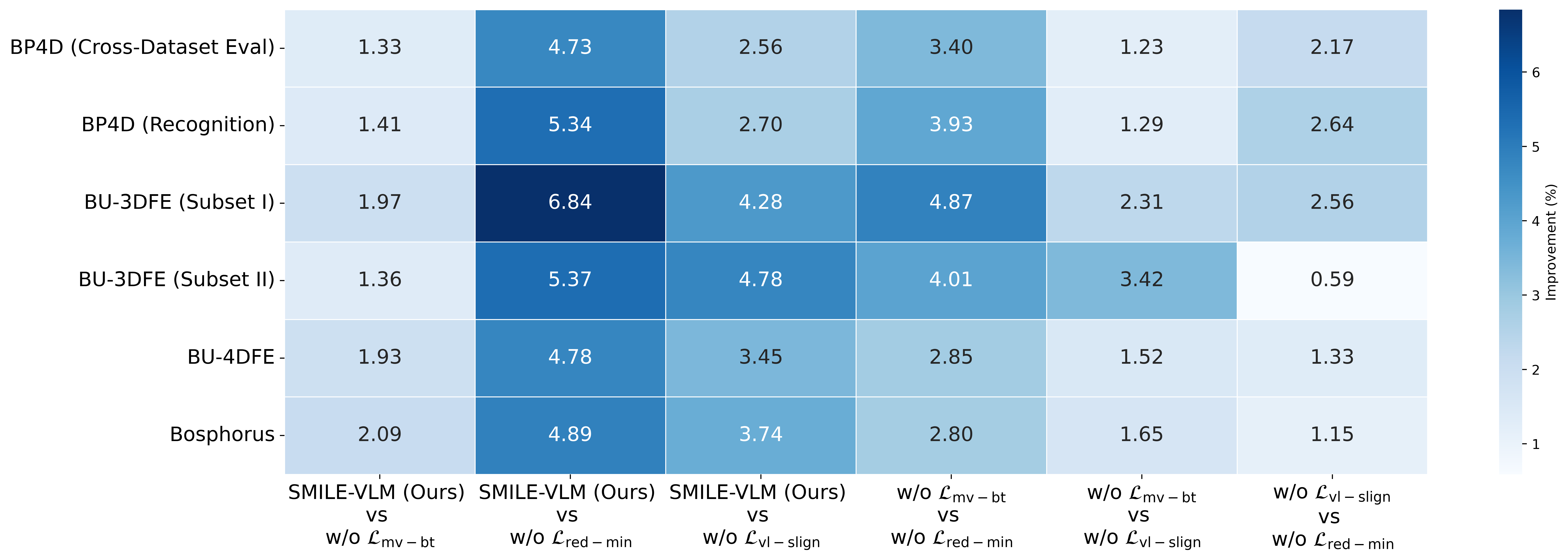}
	\caption{Heatmap illustrating accuracy improvements across multiple benchmark datasets. Each cell quantifies the gain in accuracy obtained by comparing different model configurations, with darker blue tones indicating greater improvements. The visualization highlights the effectiveness of the proposed SMILE-VLM framework.}
	\label{accuracy_imp}
\end{figure*}

In Fig.~\ref{accuracy_imp}, we present a heatmap illustrating accuracy improvements achieved through progressive integration of key components in the SMILE-VLM framework. The heatmap visualizes pairwise differences across configurations using a blue gradient, where darker shades represent larger accuracy gains. It can be noted that the most substantial improvement is observed on BU-3DFE (Subset I), where the SMILE-VLM model is ahead by a good margin of 6.84\%. Similar performance boosts are seen across BU-3DFE (Subset II), BU-4DFE, and Bosphorus, indicating consistent gains. On more challenging benchmarks like BP4D (cross-dataset evaluation), improvements are still evident, especially when \( \mathcal{L}_{\text{vl-align}} \) is included. These results validate the additive benefits of our multiview and vision-language components, highlighting that they contribute meaningfully to learning robust, expressive representations across both posed and spontaneous 3D/4D facial expression datasets.

\subsection{Extending SMILE-VLM to 4D Micro-Expression Recognition (MER)}
To further demonstrate the generalizability of SMILE-VLM, we extend our model to the task of 4D micro-expression recognition (MER) using the 4DME dataset \cite{li20224dme} and compare with their baseline results. Micro-expressions are subtle, brief facial movements that reflect underlying emotions and are often difficult to detect due to their low intensity and short duration. Given the rich spatiotemporal nature of 4D data and the semantic potential of language alignment, our proposed model is well-suited to this task. For this extension, we fine-tune our model with emotion-sensitive textual prompts designed to capture the micro-expression understanding of each class. Specifically, we augment the prompt set using templates such as \textit{``a face with [CLS] micro expression''}, where [CLS] is replaced by the emotion category, e.g., ``positive'', ``negative'', ``surprise'', ``repression'', or ``others''.

In Table~\ref{tab:4dme_mer}, we report the recognition performance of SMILE-VLM on the 4DME dataset, evaluated across left, right, and front profile views, as well as under a multi-view fusion setting. As shown in this table, our model achieves the highest average F1-score of 0.8023 and accuracy of 86.61\%, demonstrating strong capability in detecting subtle micro-expressions. The multi-view configuration significantly boosts performance across all emotion classes, especially for categories like ``repression'' and ``negative'', where fine-grained features and multi-angle cues are essential. These results validate the adaptability of SMILE-VLM to spontaneous, low-intensity facial dynamics present in MER scenarios.
\section{Conclusion}
In this work, we presented SMILE-VLM, a novel self-supervised framework for 3D/4D facial expression recognition that integrates multiview visual inputs with vision-language modeling. By leveraging redundancy reduction, cross-modal alignment, and multiview decorrelation losses, SMILE-VLM effectively learns semantically meaningful and view-invariant representations without relying on labeled emotion data. Our framework demonstrates strong generalization across multiple benchmarks, including BU-3DFE, BU-4DFE, BP4D-Spontaneous, and Bosphorus, achieving performance competitive with or superior to existing supervised and unsupervised methods. We further extended SMILE-VLM to the task of 4D micro-expression recognition to model subtle and short-lived affective cues. The model achieved high F1-scores and accuracy on the 4DME dataset, validating the adaptability of our approach to fine-grained spatial dynamics. Overall, SMILE-VLM opens new directions for scalable, label-efficient, and multimodal affective computing.


{\small
\bibliographystyle{ieeetr}
\bibliography{references}
}

\EOD

\end{document}